\documentclass[10pt,twocolumn,letterpaper]{article}
\usepackage[numbers,sort&compress]{natbib}
\usepackage{iccv}
\usepackage{times}
\usepackage{epsfig}
\usepackage{graphicx}
\usepackage{amsmath}
\usepackage{amssymb}

\usepackage{booktabs}
\usepackage{multirow}
\usepackage{enumitem}
\usepackage[hypcap=false,font=small]{caption}
\usepackage[table, svgnames]{xcolor}
\usepackage{microtype}
\usepackage{xspace}

\usepackage[pagebackref=true,breaklinks=true,letterpaper=true,colorlinks,bookmarks=false]{hyperref}

\newcommand{\backronym}{ASIC\xspace}
\newcommand{\myparagraph}[1]{\medskip\noindent\textbf{#1}}

\newcommand{\bbox}{\text{bbox}}
\newcommand{\alphapck}{\alpha_\bbox}
\newcommand{\kcycle}{\text{k-CyPCK}}
\newcommand{\cycle}{\text{-CyPCK}}

\newcommand{\I}{\mathbf{I}}
\newcommand{\Ia}{\I^\text{a}}
\newcommand{\Ib}{\I^\text{b}}

\newcommand{\F}{\mathbf{F}}
\newcommand{\Fa}{\F^\text{a}}
\newcommand{\Fb}{\F^\text{b}}
\newcommand{\f}{\mathbf{f}}
\newcommand{\fa}{\f^\text{a}}
\newcommand{\fb}{\f^\text{b}}
\newcommand{\p}{\mathbf{p}}
\newcommand{\pa}{\p^\text{a}}
\newcommand{\pb}{\p^\text{b}}
\newcommand{\A}{\boldsymbol{\Phi}_\text{align}}
\newcommand{\G}{\mathbf{G}}
\newcommand{\C}{\mathbf{C}}

\newcommand{\cc}{\mathbf{c}}
\newcommand{\cca}{\cc^\text{a}}
\newcommand{\ccb}{\cc^\text{b}}
\newcommand{\Irec}{\I_\text{Recon}}
\newcommand{\M}{\mathbf{M}}
\newcommand{\Mrec}{\M_\text{Recon}}
\newcommand{\loss}{\mathcal{L}}
\newcommand{\T}{\mathcal{T}}
\newcommand{\W}{\mathcal{W}}
\newcommand{\Id}{\mathcal{I}}

\makeatletter
\g@addto@macro{\endtabular}{\rowfont{}}
\makeatother
\newcommand{\rowfonttype}{}
\newcommand{\rowfont}[1]{
\gdef\rowfonttype{#1}#1\ignorespaces%
}
\makeatother
\usepackage[capitalize]{cleveref}
\crefname{section}{Sec.}{Secs.}
\Crefname{section}{Section}{Sections}
\Crefname{table}{Table}{Tables}
\crefname{table}{Tab.}{Tabs.}

\iccvfinalcopy 

\def\iccvPaperID{4552} 

\ificcvfinal\fi

\begin{document}

\title{\vspace{-2em} ASIC: Aligning Sparse in-the-wild Image Collections}

\author{Kamal Gupta\textsuperscript{\rm 1,2}, Varun Jampani\textsuperscript{\rm 1}, Carlos Esteves\textsuperscript{\rm 1}, \\
Abhinav Shrivastava\textsuperscript{\rm 2}, Ameesh Makadia\textsuperscript{\rm 1}, Noah Snavely\textsuperscript{\rm 1},  Abhishek Kar\textsuperscript{\rm 1}\hfill \\ \\
\textsuperscript{\rm 1}Google \quad \quad
\textsuperscript{\rm 2}University of Maryland, College Park\\
}

\twocolumn[{%
\renewcommand\twocolumn[1][]{#1}%
\maketitle
\begin{center}
    \centering
    \vspace{-1.2em}
    \includegraphics[width=0.97\textwidth]{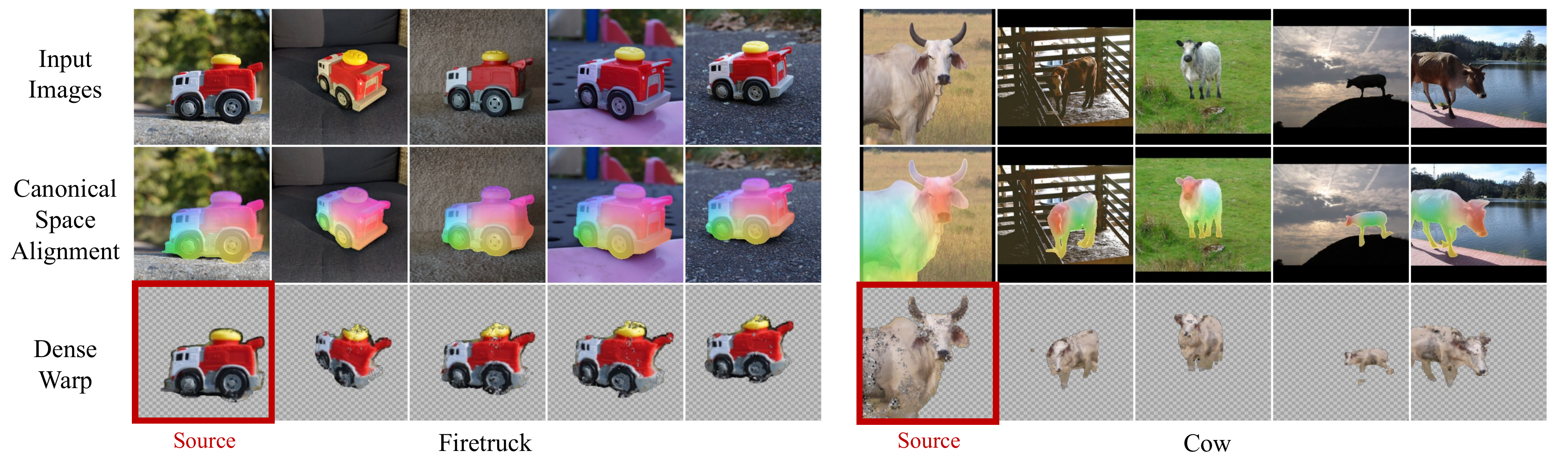}
    \vspace{-0.5em}
    \captionof{figure}{\textbf{Globally consistent and dense aligments with ASIC.} Given a small set ($\sim$10-30) of images of an object or object category captured in-the-wild, our framework computes a dense and consistent mapping between all the images in a self-supervised manner. \textbf{First row:} Unaligned sets of images from the SAMURAI (Keywest) and SPair-71k (Cow) datasets. \textbf{Second row:} Dense correspondence maps produced by our method. \textbf{Third row:} Image in the first column warped to the images in columns 2-5. 
    }
    \label{fig:teaser}
\end{center}
}] 

\maketitle
\ificcvfinal\thispagestyle{empty}\fi

\begin{abstract}
\vspace{-3mm}
  We present a method for joint alignment of sparse in-the-wild image collections of an object category. Most prior works assume either ground-truth keypoint annotations or a large dataset of images of a single object category. However, neither of the above assumptions hold true for the long-tail of the objects present in the world. We present a self-supervised technique that directly optimizes on a sparse collection of images of a particular object/object category to obtain consistent dense correspondences across the collection. We use pairwise nearest neighbors obtained from deep features of a pre-trained vision transformer (ViT) model as noisy and sparse keypoint matches and make them dense and accurate matches by optimizing a neural network that jointly maps the image collection into a learned canonical grid. Experiments on CUB and SPair-71k benchmarks demonstrate that our method can produce globally consistent and higher quality correspondences across the image collection when compared to existing self-supervised methods.
  Code and other material will be made available at \url{https://kampta.github.io/asic}.
\vspace{-5mm}
\end{abstract}

\section{Introduction}
\label{sec:intro}
\vspace{-1.5mm}

Given an image of a car, we as humans, can easily map corresponding pixels between this car and an arbitrary collection of car images. Our visual system is able to achieve this (rather impressive) feat using a multitude of cues - low level photometric consistency, high level visual grouping and our priors on cars as an object category (shape, pose, materials, illumination etc.). The above is also true for an image of a ``never-before-seen" object (as opposed to a common object category such as cars) where humans demonstrate surprisingly robust generalization despite lacking an object or category specific priors~\citep{biederman1987recognition}. These correspondences in turn inform downstream inferences about the object such as shape, affordances, and more. In this work, we tackle this problem of ``low-shot dense correspondence" -- \ie given only a small in-the-wild image collection ($\sim$10--30 images) of an object or object category, we recover dense and consistent correspondences across the entire collection.

Prior works addressing this problem of dense alignment in ``in-the-wild" image collections assume availability of annotated keypoint matches and image pairs~\citep{min2019hyperpixel,cho2021cats}, a mesh of the object~\citep{kulkarni2020articulation,zhang2021ners}, or a very large collection of images of the object~\citep{peebles2022gan,mu2022coordgan}. These assumptions often do not hold for the long-tail of objects that exist in real world imagery. This long-tail is unavoidable; no matter how many new images we annotate, we will keep uncovering new and rare categories of objects. In our work, we show that it is possible to achieve dense correspondence of small in-the-wild image collections without any manual annotations by leveraging the power of large self-supervised vision models.
Aligning these image sets can be useful for a wide range of applications such as edit propagation for images and videos, as well as downstream problems such as pose and shape estimation. 

In the presence of a limited number of samples and a high-dimensional search space, dense correspondence and joint alignment of an image set is a challenging optimization problem. We draw inspiration from classical image alignment methods~\citep{szeliski2007image,learned2005data} where images are warped (or congealed) to a consistent canonical pose before classification using simple transformations, as well as recent works on per-image-set optimization~\citep{ulyanov2018deep,mildenhall2021nerf,kasten2021layered}, where the inductive model biases coupled with additional regularization allows for learning a good solution with self-supervision. Our framework, dubbed \backronym, consists of a small image-to-image network which predicts a dense per-pixel mapping from the image to a two-dimensional canonical grid.  This canonical grid is parameterized as a multi-channel learned embedding and stores an RGB color along with an alpha value
indicating whether the location represents the object or the background. We devise a novel contrastive loss 
function to ensure that semantic keypoints from different images map to a consistent location in canonical space.

The key contribution of our work is to exploit noisy and sparse pseudo-correspondences between a pair of images and extend them to learn consistent dense correspondences across the image collection. These pseudo-correspondences can be obtained using any of the large self-supervised learning (SSL) models~\citep{chen2020simple, caron2020unsupervised, chen2020exploring, misra2020self, he2020momentum, grill2020bootstrap, radford2021learning} which learn without explicit labels on large internet-scale data. 
In order to make them accurate, we enforce pair-wise consistency across the image collection with an alignment network and a self-supervised keypoint consistency loss. Further, we introduce additional regularization via equivariance and reconstruction terms to get dense correspondences across collection. 

\cref{fig:teaser} demonstrates the dense and consistent mapping learned by our model for two image sets. We also evaluate our method on 18 image categories in SPair-71k~\citep{min2019spair} dataset, 4 categories in PF-Willow~\citep{ham2016}, 3 fine-grained categories in the CUB~\citep{wah2011caltech}, as well as 5 collections in SAMURAI~\citep{boss2022samurai} datasets and show that \backronym is competitive against unsupervised keypoint correspondence approaches, and often outperforms them. An additional advantage of learning a joint canonical mapping is that our method suffers significantly less drift when propagating keypoints on a sequence of images (instead of just a single image pair). In order to evaluate the keypoint consistency over a sequence of $k$ images, we propose a new metric $\kcycle$ (or k-cycle PCK) in \cref{sec:quant_consistent}
and show that our method outperforms existing methods by over 20\% at both the low and high precision settings. In summary, our contributions are as follows:
\begin{itemize}[leftmargin=*,itemsep=0em]
\item We introduce a test-time optimization technique to recover consistent dense correspondence maps over a small collection of in-the-wild images.
\item We design a novel loss function and several regularization terms to encourage mapping to be consistent across multiple images in a given collection.
\item We perform extensive quantitative and qualitative evaluations on 4 different datasets (spanning 30 object categories) to show that our method is competitive with the unsupervised methods, often outperforming them.
\item We propose a novel metric $\kcycle$ to evaluate consistency of keypoint propagation over a sequence of images, which is not captured by traditional metrics such as PCK.
\end{itemize}

\section{Related Work}
\label{sec:related}

\noindent \textbf{Correspondence between image pairs.}
Keypoint matching or correspondence between images is one of the oldest 
tasks in computer vision. Some very early works focused on finding dense optical flow~\citep{horn1981determining,black1993framework,beauchemin1995computation} between pairs of consecutive images in videos via a variational framework to optimize flow based on pixel intensities. 
Sparse keypoint matching, e.g., using SIFT descriptors~\citep{lowe2004sift}, also gained importance due to applications in tracking~\citep{lucas1981iterative,tomasi1991detection} and structure from motion (SfM)~\citep{frahm2010building,agarwal2011building,schonberger2016structure}. 
SIFT Flow~\citep{liu2010sift} proposed the idea of using SIFT descriptors for dense alignment between image pairs. 
Initial deep learning based correspondence works~\citep{choy2016universal,kim2017fcss,detone2018superpoint} replaced SIFT with deep features. With the availability of labeled datasets, a number of works have performed end-to-end matching with deep networks~\citep{min2019hyperpixel,liu2020semantic,li2021probabilistic,jiang2021cotr,sun2021loftr,sarlin2020superglue,rocco2018neighbourhood,li2020correspondence,zhao2021multi,min2021convolutional,lee2021patchmatch,huang2022learning}. However, a shortcoming of these aforementioned works is that they usually require large labeled datasets, and often fail to generalize on unseen objects or scenes.

\myparagraph{Joint alignment of image sets.}
The concept of a canonical image has long been used for the task of object detection via template matching~\citep{gavrila1998multi,ioffe2001probabilistic}. 
Learned-Miller~\etal~\citep{huang2007unsupervised,learned2005data} formalized the task of jointly aligning a set of images (i.e., congealing them) by continuously warping each image (\eg via affine transformations) to minimize the entropy distribution of the image set. 
\cite{huang2012learning} use deep features from multiple resolutions in place of hand-crafted features. 
GANgealing~\citep{peebles2022gan} extended this idea by constraining the canonical image to be the output of a pre-trained StyleGAN~\citep{karras2021alias,goodfellow2020generative}. 
In a similar vein, CoordGAN~\citep{mu2022coordgan} trains a structure-texture disentangled GAN with a canonical coordinate frame as input. 
Both of these works attempt to solve a similar tasks as ours, but are limited by data-hungry GAN training. Some works exploits 3D shape as a means for consistent dense correspondences across image collections~\citep{kulkarni2020articulation, kulkarni2019csm, cmrKanazawa18,yao2022lassie} but require access to additional signals such as category specific 3D templates, segmentation masks or keypoint correspondences. In contrast, our work attempts to learn dense correspondences in a low-shot setting where GAN training is infeasible and in the absence of additional training signals. As mentioned before, we do so primarily by leveraging large pre-trained SSL models as our source of semantic priors on general imagery.

\myparagraph{Self-supervised correspondence discovery.}
To overcome the lack of large datasets with ground-truth correspondence, recent work seeks to combine the idea of distilling deep features from a network trained with self-supervision on large-scale image datasets. Some of these works optimize for proxy losses computed with known  transformations~\citep{seo2018attentive, novotny2018self, truong2021warp,thewlis2017unsupervised,aygun2022demystifying,thewlis2019unsupervised,jeon2018parn,rocco2018end,kim2019semantic}. Like these methods, we also train our network to be equivariant to synthetic geometric transformations. However, a key difference is that we also train with pseudo-correspondences `across' real images, which allows the method to generalize better and build a consistent mapping across the given image collection.

Deep Matching Prior ~\citep{hong2021deep}
and Neural Best Buddies~\citep{aberman2018neural} optimize for only a single pair of images to match deep features of one image to another. More recently PSCNet~\cite{jeon2021pyramidal} and Neural Congealing~\citep{ofri2023neural} train large networks for simultaneously matching deep features for image pairs by learning a flow from image to image, and image to canonical space respectively. However, these methods have limited flexibility in the deformation space and do not generalize well to out of plane rotations present in datasets such as SPair-71k. We allow our model to map different image regions arbitrarily to different parts of the canonical space. In \cref{tab:spair_pck}, we show that this allows us to generate more accurate correspondences and generalize to more object categories.

\section{\backronym~Framework}

\begin{figure*}
 \centering
 \includegraphics[width=\linewidth,trim={0 0 0 0}, clip]{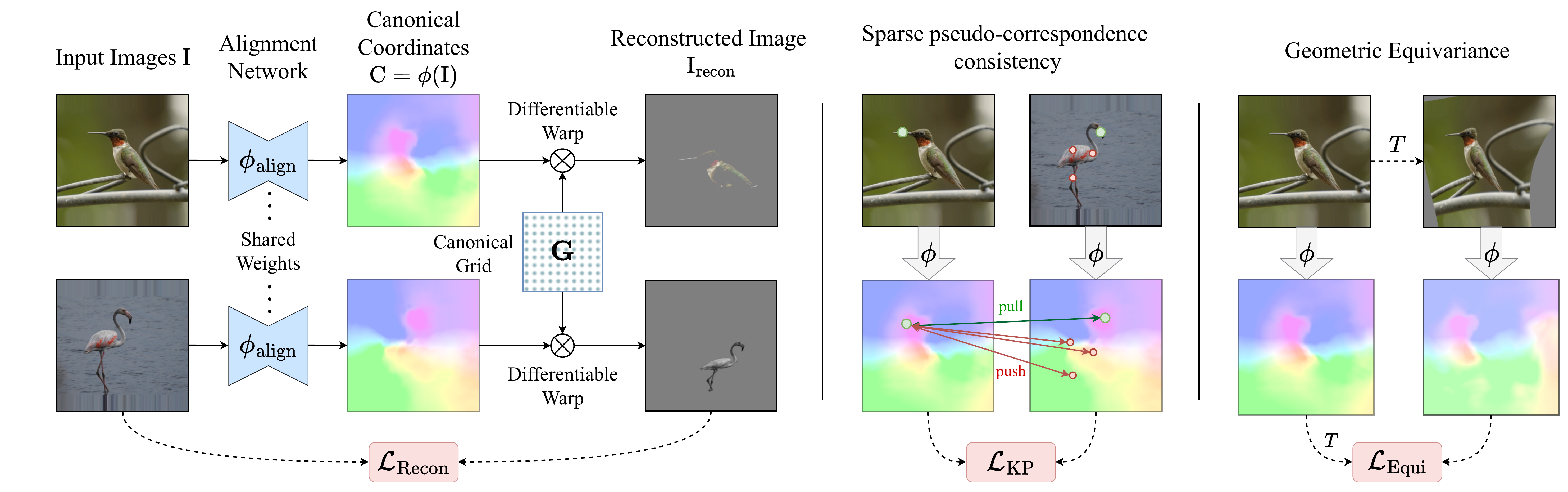}
 \caption{\footnotesize \textbf{\backronym Architecure.} The alignment network $\A$ predicts canonical space coordinates for all pixels for all input images. Images can be reconstructed using a differentiable warp from the canonical space. In order to align semantically similar pixels from different images to the same location in the canonical space, we propose two primary loss functions $\loss_\text{KP}$ and $\loss_\text{Recon}$. Please refer to the \cref{sec:arch} for more details.}
 \label{fig:arch}
 \vspace{-0.5cm}
\end{figure*}

Given a collection of images of an object or an object category, our goal is to assign corresponding pixels in all the images to a unique location in a canonical space. By doing so, we can use this learned canonical space as an intermediary when mapping pixels from one image to any another image in the collection while guaranteeing global consistency. The absence of ground truth annotations, small size of the datasets we consider ($\sim$10 - 30 images) and the presence of occlusions and variations in shape, texture, viewpoint, and background lighting all serve to make this task highly challenging. We introduce a simple yet robust framework with a novel self-supervised contrastive loss function over image pairs, as well as auxiliary regularization losses on this learned canonical space to find consistent dense correspondences across the collection.

\subsection{Obtaining Pseudo-correspondences}
\label{sec:preprocessing}

Prior work has shown that deep features extracted from large pre-trained networks contain useful local semantic information~\citep{caron2020unsupervised,choudhury2021unsupervised,hung2019scops}. In this work, we use DINO~\citep{caron2020unsupervised} to extract these local semantic features. Note that these features are only extracted for obtaining pseudo-correspondences only once and are not used during the training.
Given a pair of images $\Ia$ and $\Ib$, we obtain feature maps $\Fa$ and $\Fb$ using DINO. Here $\Fa = \{\fa_i\}$ and $\Fb = \{\fb_i\}$ represents the sets of feature vectors $\f \in \mathbb{R}^\text{d}$ for all spatial locations $\p_i=(x,y) \in \mathbb{R}^2$. In practice, we obtain these feature maps at a coarser resolution, but for brevity, we do not introduce new notations for low-resolution feature maps. 
We define our pseudo keypoint correspondences, between the two images $\Ia$ and $\Ib$ as all pairs of locations of feature vectors that are mutual nearest neighbors, \ie,
\begin{equation*}
    \{(\pa_i, \pb_j)\ |\ \big(\text{NN}(\fa_i, \Fb)=\fb_j\big) \ \wedge\  \big(\text{NN}(\fb_j, \Fa)=\fa_i\big)\}
\end{equation*}
where $\text{NN}(\fa_i, \Fb)$ corresponds to the nearest neighbor of the normalized feature vector $\fa_i$ in the set of feature vectors $\Fb$.
The mutual nearest neighbors are usually noisy and sparse, and they serve as pseudo-correspondences for training our alignment network which we discuss next.

\subsection{Architecture}
\label{sec:arch}
\cref{fig:arch} gives the high-level overview of the framework. Formally, we are given an image collection consisting of $N$ images $\{\I^k\}_{k=1}^N$. We want to train an alignment network $\A$ that 
takes a single image as input at a time and outputs $\C = \A(\I)$, the canonical space coordinate map, of the image. The canonical space coordinate map $\C$ has the same spatial dimensions as the input $H \times W$ and contains $(u,v)$ coordinates in the shared canonical grid for that location. We parameterize this alignment network $\A: \mathbb{R}^{H \times W \times 3} \rightarrow \mathbb{R}^{H \times W \times 2}$ with a fully convolutional U-Net~\citep{ronneberger2015u} trained from scratch for the collection. Each pixel location $\p = (x, y)$ of this map consists of a 2-dimensional $\cc = (u, v)$ coordinate.

These coordinates corresponds to a location in a learned canonical grid $\G \in \mathbb{R}^{H' \times W' \times 4}$. The canonical grid $\G$ is also two-dimensional but can have arbitrary height and width $H' \times W'$, and is shared by all the images in the collection.
Each location in canonical grid $\G$ stores an $(r,g,b,\alpha)$ value which corresponds to colors $(r,g,b)$ and a probability $\alpha$ that this location corresponds to a foreground pixel in the image. The original image, and a foreground visibility mask can now be reconstructed using this shared canonical grid $\G$, canonical space mapping $\C$, and a differentiable warp operator commonly used in spatial transformer networks~\citep{jaderberg2015spatial}.
For the mapping to be meaningful, we want semantically similar points from different images to map to the same location in the canonical space. In the next section, we describe the training loss we devised to this end. 

\subsection{Training Objectives}
\label{sec:training}
\noindent\textbf{Sparse pseudo-correspondence consistency.} The central goal of our framework is to ensure that semantically similar points in the images are aligned in the canonical space.  Recall from the \cref{sec:preprocessing}, that we pre-compute the pseudo-correspondences between all pairs of images using mutual nearest neighbors in the SSL feature space. Since SSL models are not trained for the task of correspondence, the pseudo-correspondences are noisy and sparse. Our first loss term is targeted at improving the accuracy of the correspondences by jointly aligning them for all pairwise combinations of images in our collection.
Formally, given an image pair $(\Ia, \Ib)$, we denote all the $K$ pseudo-correspondences in the pair by $\{\pa_i, \pb_i\}_{i=1}^K$. We apply the alignment network $\A$ to $\Ia$ and $\Ib$ independently to obtain the canonical space coordinates for each pixel in the pair, which we denote by $\{\cca_i, \ccb_i\}_{i=1}^K$. We want to map each keypoint location in $\Ia$ as close as possible to its counterpart in $\Ib$, while pushing it away from the mapping of other keypoints in $\Ib$. To achieve this, we define our first loss function $\loss_\text{KP}$ as
\begin{align}
    \loss_\text{KP} &= -\sum_{i=1}^{K}{\log{\frac{\exp(-\parallel\cca_i - \ccb_i\parallel^2/\tau)}{\sum_{j=1}^{K}{\exp(-\parallel\cca_i - \ccb_j\parallel^2}/\tau)}}}
\end{align}
where $\tau$ is a hyperparameter and is fixed to $1.0$ in all our experiments. $\loss_\text{KP}$ plays the key role in improving the accuracy of pseudo-correspondences jointly for all images in our collection. However, the number of pseudo-correspondences is still very small (typically 100-300 for an image pair) as compared to the number of pixel locations. Hence, this loss is sparse and we need to add extra regularization terms in order to learn dense alignment, that we will discuss next.

\myparagraph{Geometric transformation equivariance.}
In order to make our learned mapping dense, we introduce a geometric equivariance regularization term in our loss function. We apply a random synthetic geometric transformation $\T$ to a given image $\I$. Since the output of the alignment network $\A$ learns the canonical space coordinates for each location of input image, we can apply the same geometric transformation $\T$ to $\A(\I)$,  and enforce an equivariance loss as follows
\begin{align}
    \loss_\text{Equi} &= \parallel\T(\A(\I)) - \A(\T(\I))\parallel
\end{align}
where $\T$ is the geometric transformation. We choose thin plate spline (TPS) transformations~\citep{duchon1977splines} in our work, commonly used for image warps. $\loss_\text{KP}$ and $\loss_\text{Equi}$ serve as the two primary loss functions for the image set alignment problem, serving the purpose of making the pseudo-correspondences accurate and dense respectively. To further aid the training, we also propose the following auxiliary regularizations.

\myparagraph{Total variation regularization.}
In order to encourage smooth mappings from from each image to the canonical space, we add a total variation (TV) regularization to the computed mapping $\C$. We found TV loss to be crucial to mitigate degenerate solutions (see \cref{sec:ablation}):
\begin{gather}
   \loss_\text{TV} = \loss_\text{Huber}(\Delta_x (\C-\Id)) + \loss_\text{Huber}(\Delta_y (\C-\Id))
\end{gather}
where $\Id$ is the identity mapping (\ie each pixel $(x, y)$ in the image gets mapped to $(x,y)$ in the canonical space), $\Delta_x$ and $\Delta_y$ denote the partial derivatives under finite differences \wrt $x$ and $y$ dimensions, and $\loss_\text{Huber}$ denotes the Huber loss~\cite{huber1992robust}.

\begin{figure*}[!ht]
  \centering
  \vspace{-2em}
  \includegraphics[width=\linewidth]{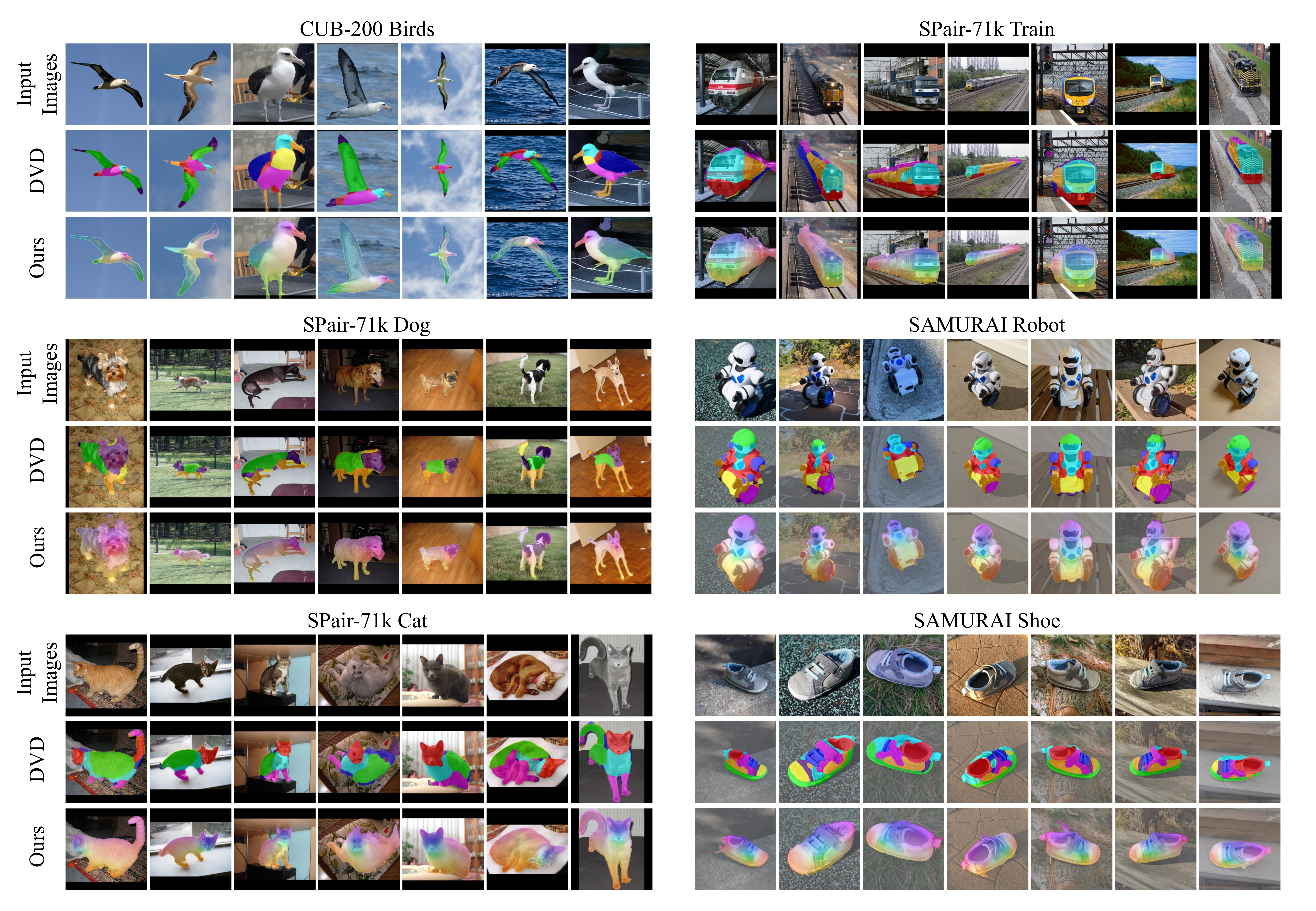}
  \vspace{-2.5em}
  \caption{\textbf{Visualizing canonical space alignment.} 
  For each dataset, the top row shows sample images from the dataset (composed of 10-30 images each). 
  The middle row shows part co-segmentations computed by DVD~\citep{amir2021deep}. DVD computes a coarse, discrete set of parts across the dataset. 
  The bottom row shows the continuous canonical space mapping computed by our method. Our canonical space mapping is 
  smooth and consistent across the images for each dataset/collection.}
  \vspace{-1em}
  \label{fig:canon}
\end{figure*}

\myparagraph{Reconstruction loss.} All the loss terms so far are computed on the canonical coordinates given by $\A$ and does not use the canonical grid $\G$. Recall that $\G$ is of the size  $H' \times W' \times 4$, and allows us to reconstruct each image as well as a foreground visibility mask via a differentiable warp operator ($\W$)~\citep{jaderberg2015spatial} such that $\Irec, \Mrec=\W(\G, \C)$.
This allows us to compute a per image reconstruction loss using the original and reconstructed images. However a simple $L_1$ or $L_2$ loss will not suffice since $\G$ is shared for all the images in the collection and these images may come from wildly different backgrounds and lighting conditions.
Furthermore, the two images might contain two different instances of the same object class, and may have different textures, shapes, and viewpoints.
We instead minimize the perceptual (LPIPS) loss~\citep{zhang2018unreasonable} which measures a perceptual patch level similarity between images.
For the reconstructed mask, we compute pixel-wise binary cross entropy (BCE) loss using the image foregrounds obtained with co-segmentation~\cite{amir2021deep}.
\begin{gather}
   \Irec, \Mrec = \W(\G, \C) \nonumber \\
   \loss_\text{Recon} = \text{LPIPS}(\I, \Irec) + \text{BCE}(\M, \Mrec)
\end{gather}

\noindent\textbf{Consistent part alignment.}
For our final auxiliary loss, we obtain  part co-segmentation maps from the images~\citep{amir2021deep,choudhury2021unsupervised} by clustering deep ViT features into $S$ semantic parts, then running GrabCut~\citep{rother2004grabcut} to smoothen the part boundaries. Our hypothesis is that semantically similar parts in the images should get mapped to similar location in $\G$, 
Formally, for each image $\I \in \mathbb{R}^{H\times W\times 3}$, we obtain semantic part masks as a binary matrix, $\textbf{P} \in \mathbb{R}^{H\times W \times S}$. Since we want a part across the image set to map to a compact location in the canonical space, we minimize the variance of the canonical space coordinates for all pixels belonging to a part:
\begin{align}
\label{eq:parts}
    \loss_\text{Parts} &= \sum_{s=1}^S \frac{1}{N_s}\sum_i \parallel \cc_i^s - \mathbb{E}(\cc^s) \parallel^2
\end{align}
where $N_s$ is the number of pixels belonging to the part, $\cc_i^s$ is the canonical coordinates of $i^\text{th}$ pixel location belonging to the $s^\text{th}$ part, and $\mathbb{E}(\cc_i^s)$ is the centroid of the $s^\text{th}$ part.
We fix the number of parts to 8 in all our experiments. Alternately, the number of parts can be computed using the elbow method~\citep{thorndike1953belongs} at the expense of additional compute.

\begin{figure*}[!ht]
  \centering
  \includegraphics[width=\linewidth]{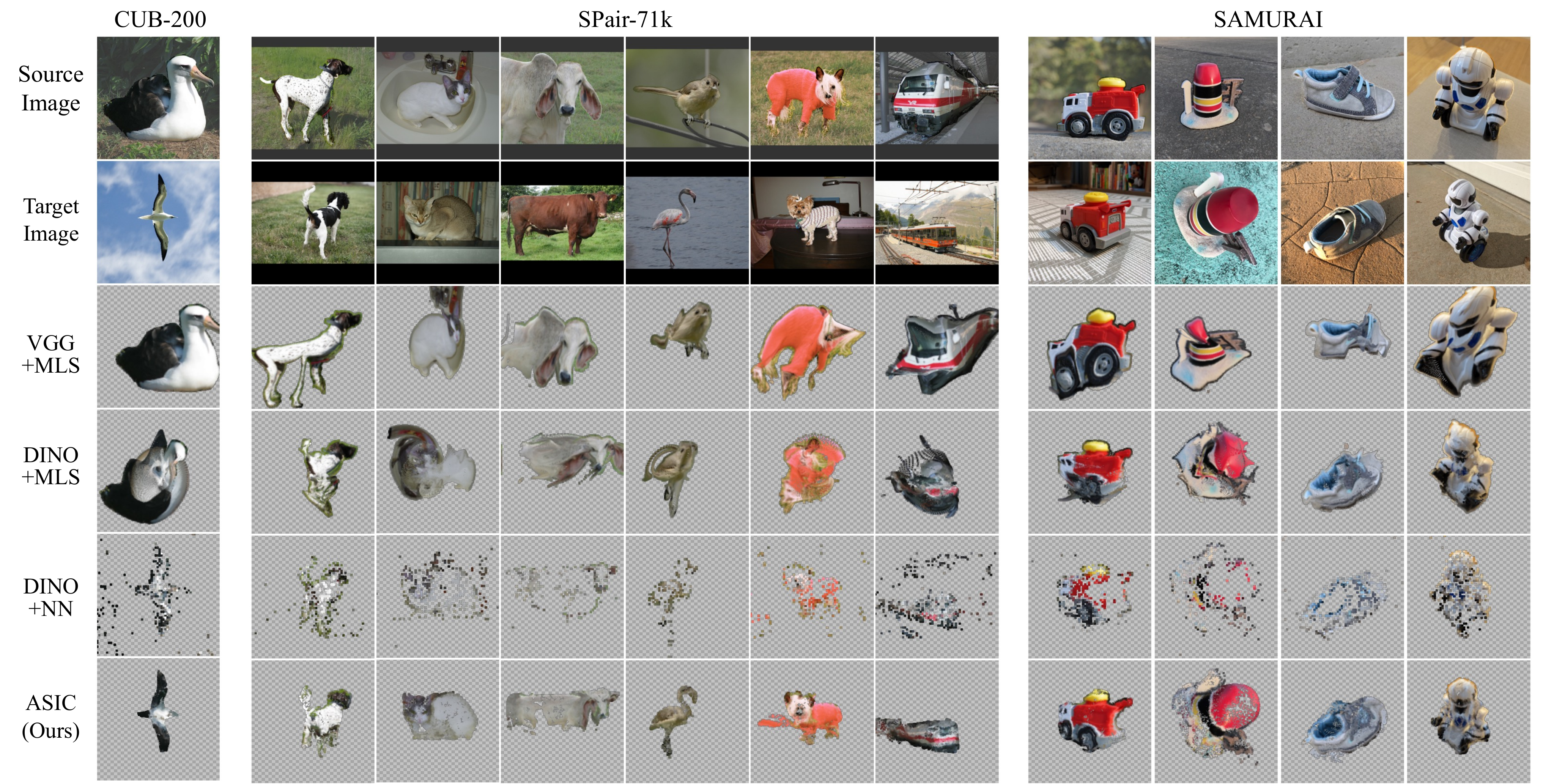}
  \caption{\textbf{Dense warping} from a source image (top row) to a target image (second row). We warp all foreground pixels (highlighted by a red overlay in the source image). Our methods produce dense and semantically more meaningful warps from the source to the target.}
  \label{fig:warp}
  \vspace{-1em}
\end{figure*}

\section{Experiments}
\label{sec:experiments}

We evaluate our method on several real-world in-the-wild image collections of both rigid and non-rigid object categories. For all datasets, we use a fixed set of hyperparameters (provided in the appendix) unless specified otherwise. 


\myparagraph{Datasets.}
\textbf{SPair-71k}~\citep{min2019spair} consists of 1,800 images from 18 
categories. We optimize over image collections derived from the SPair-71k test set for each category independently and report results
 on each individual category, as well as aggregate results over all 18 categories. In case of \textbf{PF-Willow}~\citep{ham2016}, we consider all 4 categories of the dataset containing $\sim$30 images. 
\textbf{CUB-200}~\citep{wah2011caltech} datasets consists of over 200 fine-grained categories. We optimized our model on the test sets of first 3 categories of the dataset, consisting of 15-20 images each.
We also show qualitative results on 4 objects from \textbf{SAMURAI} dataset~\citep{boss2022samurai}.


\subsection{Canonical Space Alignment}
\label{sec:qual_consistent}

One simple way to visualize the alignment of an image set when mapped to the canonical space  $\G$
is to define a colormap over the canonical grid and color the image pixels according to their mapped location in the canonical grid. 
In \cref{fig:canon}, the first row for each collection contains sample input images. The second row shows discrete parts obtained via parts co-segmentation using \citep{amir2021deep}. 
While these parts are also consistent across the image set, our canonical space mapping (third row) can be seen as a dense and continuous co-segmentation. We show the results for six datasets: CUB-200 birds; Dogs, Cats, and Train from SPair-71k; and Robot and Shoe from SAMURAI. 
The colormap used for the canonical space is provided in the supplement.
We observe that our method can find dense correspondences across highly varying poses, backgrounds, and lighting.
It also maps common parts of objects in a dataset to nearby regions of the canonical space.
This is evident in \cref{fig:canon} where, for instance, the faces of different cats are colored similarly.

\subsection{Visualizing Dense Correspondences}
\label{sec:qual_dense}
We can also find dense correspondences between a pair of images $\Ia$ and $\Ib$ using our framework. Recall that $\A$ outputs canonical space coordinates $\cca$ and $\ccb$ for each pixel location $\pa$ and $\pb$.
In order to warp the source image $\Ia$ to a target image $\Ib$, for every foreground pixel $\pa_i$ in $\Ia$, we need to find its nearest neighbor among the set of points in $\pb$ in the canonical space:
We perform this action for all the foreground pixels in the source image, and splat according to the nearest neighbor mapping to get our desired warped image. 
\cref{fig:warp} shows qualitative results 
for 10 different datasets. The top row is the source image with foreground mask highlighted. The second row is the target image. We show results for two other pairwise image optimization approaches, and then our method in the last row. NBB~\citep{aberman2018neural} computes nearest neighbors using VGG-19 and applies a Moving Least Squares (MLS) optimization~\citep{schaefer2006image} to compute a dense flow from source to target. While the flow computed via MLS is smooth, it usually does not respect semantic correspondences, as evident in the figure. 
We extend their technique to use DINO features as well.
With DINO and the nearest neighbor approach (DVD)~\citep{amir2021deep}, the semantic correspondences are arguably better, but since this approach relies on the output of a Vision Transformer (ViT), which has lower resolution than the image, it produces a sparse flow. Our method produces both dense and consistent flow between an image pair.

\begin{table*}[t]
\centering
\caption{\textbf{Evaluation on SPair-71k.} Per-class and average PCK@$0.10$ on test split. Highest PCK among \textit{weakly supervised} methods in bold, second highest underlined. Scores marked with ($\star$) means the paper uses a fixed image from the test set as canonical image. Our method is competitive against other weak supervised approaches and often outperforms them.}
\label{tab:spair_pck}
\resizebox{\textwidth}{!}{
\begin{tabular}{@{}clccccccccccccccccccc@{}}
\toprule
\textbf{Supervision} & \textbf{Method}  & Aero & Bike & Bird & Boat &Bottle& Bus  & Car  & Cat  & Chair& Cow  & Dog  & Horse& Motor&Person& Plant& Sheep & Train & TV & \textbf{All}\\  
\midrule
\multirow{1}{3.2cm}{\centering{Strong Supervision}}
& SCorrSAN~\citep{huang2022learning}    & 57.1 & 40.3 & 78.3 & 38.1 & 51.8 & 57.8 & 47.1 & 67.9 & 25.2 & 71.3 & 63.9 & 49.3 & 45.3 & 49.8 & 48.8 & 40.3 & 77.7 & 69.7 & 55.3\\
\rowfont{\color{Black}}
\multirow{1}{3.2cm}{\centering{GAN supervision}}
& GANgealing~\citep{peebles2022gan}     &    - & 37.5 &    - &    - &    - &    - &    - & 67.0 &    - &    - & 23.1 &    - &    - &    - &    - &    - &    - & 57.9 &    -\\
\hline
\multirow{7}{3.2cm}{\centering{Weak supervision \\ (train/test)}}
& CNNGeo~\citep{rocco2017convolutional} & 23.4 & 16.7 & 40.2 & 14.3 & 36.4 & 27.7 & 26.0 & 32.7 & 12.7 & 27.4 & 22.8 & 13.7 & 20.9 & 21.0 & 17.5 & 10.2 & 30.8 & 34.1 & 20.6\\
& A2Net~\citep{seo2018attentive}        & 22.6 & 18.5 & 42.0 & 16.4 & 37.9 & \textbf{30.8} & 26.5 & 35.6 & 13.3 & 29.6 & 24.3 & 16.0 & 21.6 & 22.8 & \textbf{20.5} & 13.5 & 31.4 & 36.5 & 22.3\\
& WeakAlign~\citep{rocco2018end}        & 22.2 & 17.6 & 41.9 & 15.1 & 38.1 & 27.4 & 27.2 & 31.8 & 12.8 & 26.8 & 22.6 & 14.2 & 20.0 & 22.2 & 17.9 & 10.4 & 32.2 & 35.1 & 20.9\\
& NCNet~\citep{rocco2018neighbourhood}  & 17.9 & 12.2 & 32.1 & 11.7 & 29.0 & 19.9 & 16.1 & 39.2 &  9.9 & 23.9 & 18.8 & 15.7 & 17.4 & 15.9 & 14.8 &  9.6 & 24.2 & 31.1 & 20.1\\
& SFNet~\citep{lee2019sfnet}            & 26.9 & 17.2 & 45.5 & 14.7 &  \underline{38.0} & 22.2 & 16.4 & \textbf{55.3} & 13.5 & 33.4 & 27.5 & 17.7 & 20.8 & 21.1 & 16.6 & 15.6 & 32.2 & 35.9 & 26.3 \\
& PMD~\citep{li2021probabilistic}       & 26.2 & 18.5 & 48.6 & 15.3 & \textbf{38.0} & 21.7 & 17.3 & 51.6 & 13.7 & 34.3 & 25.4 & 18.0 & 20.0 & 24.9 & 15.7 & 16.3 & 31.4 & \textbf{38.1} & 26.5\\
& PSCNet-SE~\citep{jeon2021pyramidal}   & 28.3 & 17.7 & 45.1 & 15.1 & 37.5 &  \underline{30.1} & \underline{27.5} & 47.4 & 14.6 & 32.5 & 26.4 & 17.7 & 24.9 & 24.5 & \underline{19.9} & 16.9 & 34.2 & \underline{37.9} & 27.0\\
\hline
\multirow{5}{3.2cm}{\centering{Weak supervision \\ (test-time optimization)}}
& VGG+MLS~\citep{aberman2018neural}     & 29.5 & 22.7 & 61.9 &  \textbf{26.5} & 20.6 & 25.4 & 14.1 & 23.7 & 14.2 & 27.6 & 30.0 & 29.1	& \underline{24.7} & 27.4 &	19.1 & 19.3 & 24.4 & 22.6 & 27.4\\
& DINO+MLS~\citep{aberman2018neural,caron2020unsupervised}& 49.7 & 20.9 & 63.9 & 19.1 & 32.5 & 27.6 & 22.4 & 48.9 & 14.0 & 36.9 & 39.0 & \underline{30.1} & 21.7 & \underline{41.1} & 17.1 & 18.1 & \underline{35.9} & 21.4  & 31.1\\
& DINO+NN~\citep{amir2021deep}          & \underline{57.2} &  24.1 &  \underline{67.4} & 24.5 & 26.8 & 29.0 & 27.1 & 52.1 &  \underline{15.7} &  \underline{42.4} &  \underline{43.3} & 30.1  & 23.2 & 40.7 & 16.6 & \underline{24.1} & 31.0 & 24.9 & \underline{33.3}\\
& NeuCongeal~\citep{ofri2023neural}& -    & \textbf{29.1$^\star$}    &    - &    - &    - &    - &    - & 53.3    &    - &    - & 35.2   &    - &    - &    - &    - &    - &    - & -    &    -\\
& \backronym~(Ours)                    & \textbf{57.9} & \underline{25.2} & \textbf{68.1} &  \underline{24.7}	& 35.4 & 28.4 & \textbf{30.9} &  \underline{54.8}	& \textbf{21.6} & \textbf{45.0} &	\textbf{47.2} & \textbf{39.9} & \textbf{26.2} & \textbf{48.8} &	14.5 & \textbf{24.5} & \textbf{49.0} & 24.6 & \textbf{36.9}\\

\bottomrule
\end{tabular}}
\vspace{-1em}
\end{table*}


\begin{table}[!ht]
\centering
    \caption{\textbf{CUB-200 and PF-Willow.} PCK@$0.10$ for three CUB categories and four PF-Willow categories.}
    \label{tab:cub_pck}
    \resizebox{\linewidth}{!}{
    \begin{tabular}{ lcc|cc } 
    \toprule
     & \multicolumn{2}{c|}{CUB-200 (3 categories)} & \multicolumn{2}{c}{PF-Willow (4 categories)} \\
     \midrule
    \textbf{Method} & PCK@$0.1$ & PCK@$0.05$ & PCK@$0.1$ & PCK@$0.05$  \\
    \midrule
    PMD~\citep{li2021probabilistic}        & - & - & 74.7 & 40.3 \\
    PSCNet-SE~\citep{jeon2021pyramidal}    & - & - & 75.1 & 42.6  \\
    VGG+MLS~\citep{aberman2018neural}      & 25.8 & 18.3 & 63.2 & 41.2   \\
    DINO+MLS~\citep{aberman2018neural,caron2020unsupervised} & 67.0 & 52.0 & 66.5 & 45.0 \\
    DINO+NN~\citep{amir2021deep}               & 68.3 & 52.8 & 60.1 & 40.1 \\
    \backronym~(Ours)                      & \textbf{75.9} & \textbf{57.9} & \textbf{76.3} & \textbf{53.0} \\
    \bottomrule
    \end{tabular}
    }
    \vspace{-1.2em}
\end{table}

\subsection{Pairwise Correspondence}
\label{sec:quant_pair}
\myparagraph{Metric.}
For evaluating accuracy of pairwise correspondence, we use the PCK metric~\citep{yang2012articulated} (percentage of correct keypoints) on the SPair, CUB, and PF-Willow datasets.


\myparagraph{Baselines.}
We categorize prior works based on the supervision used: 
(1) \emph{Strong supervision} methods utilize human-annotated keypoints to learn pairwise image correspondence and achieve the best performance (on average). We include the numbers from a recent work~\citep{huang2022learning} for reference purposes.
(2) \emph{GAN supervision} methods like~\citep{peebles2022gan} use a category-specific GAN pre-trained with large external datasets. While this method works well, it is restricted to only the categories for which large datasets are available and GAN training is feasible.
(3) \emph{Weak supervision} methods use category-level supervision (\ie they assume that given pair/collection of images are from same category). They often resort to fine-tuning a large ImageNet~\citep{deng2009imagenet} pre-trained network using a self-supervised loss function (\eg with synthetic transformations) and optionally use additional information such as foreground masks or matching image pairs for training.
Some of these works follow a \emph{train/test} setting, where the network is fine-tuned on a separate set of training images.  Note that in our work, we train a much smaller network from scratch instead of fine-tuning a large network. Some approaches (including ours) directly perform \emph{test-time optimization} without additional training data or annotations.


NBB \cite{aberman2018neural} optimizes a flow from one image to another using mutual nearest neighbors as control points~\citep{schaefer2006image}. While~\cite{aberman2018neural} shows the results by computing nearest neighbors from a VGG network, we further extend their work to utilize a DINO network. \cite{amir2021deep} simply computes nearest neighbors in DINO feature space. 
A concurrent work, Neural Congealing~\cite{ofri2023neural}, is closest to our work, in that they also perform test-time training using a canonical atlas. 
However, for objects with large deformations (such as in SPair-71k), they need to apply category specific accommodations (for instance, fixing the atlas for bicycle category). 
Our canonical grid allows for large deformations and is learned in all cases with a fixed set of hyperparameters.
We obtain scores for other models from their respective papers (whenever available) or from~\citep{huang2022learning}. 
Scores for~\cite{aberman2018neural,amir2021deep,caron2020unsupervised} are computed using official code. 
The official code of~\cite{ofri2023neural} did not converge on several objects in our experiments, hence we report the quantitative results from the paper.

\myparagraph{Discussion.} \cref{tab:spair_pck} shows PCK@$0.1$ for all SPair-71k~\citep{min2019spair} categories.
It is evident that having groundtruth keypoint annotations
during training is highly beneficial; approaches that lack keypoints 
during training lag behind.
We also observe that for categories with rigid objects (or less extreme deformations) such as `Bottle' or `Bus', weakly supervised approaches attain a similar performance as ours. However, in the objects with extreme variations such as animals/birds, our method outperforms other baselines. In our experiments, we observed that per-category hyperparameters can increase PCK performance further by $\sim 2\%$. This strategy is similar to Neural Congealing~\cite{ofri2023neural} where specific accommodations are made per category (\eg tailored training regime for bicycle). However we report our numbers with a fixed set of hyperparameters for consistency.

\cref{tab:cub_pck} shows average results for the first 3 categories of the CUB dataset, and 4 categories of the PF-Willow dataset. Note that PF-Willow is 
an easier dataset
compared to SPair-71k since it consists of rigid objects with little variation. 
Our method has performance similar to PSCNet-SE~\cite{jeon2021pyramidal} when we compute PCK using threshold $\alphapck=0.1$ (which corresponds to a $\sim$20-pixels margin of error). However at higher precision ($\alphapck=0.05$), our method provides much larger gains compared to the baselines.




\begin{figure}[t]
  \centering
  \includegraphics[width=\linewidth]{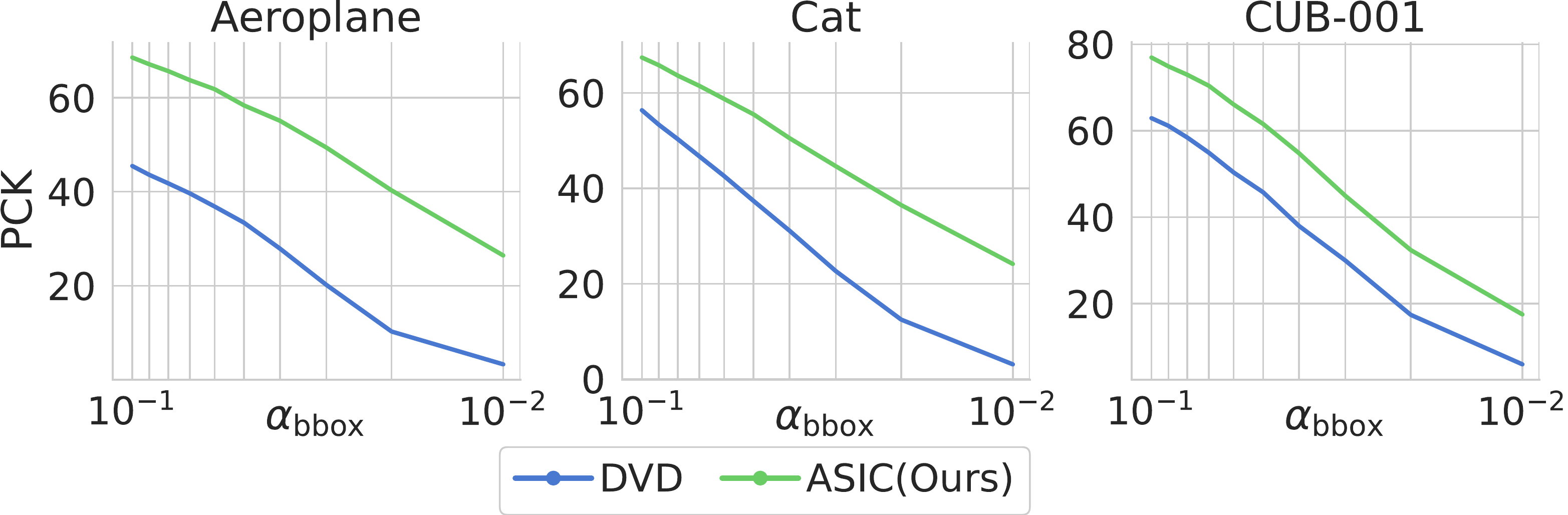}
  \caption{\footnotesize \textbf{Image Set Correspondence.} $\kcycle$ at varying $\alphapck$ (higher is better). Our method outperforms DVD baseline at both large and small values of $\alphapck$.}
  \label{fig:cycle_pck}
\end{figure}

\subsection{Image Set Correspondence}
\label{sec:quant_consistent}
Our goal in this work is to recover dense and \textit{consistent} correspondences. A shortcoming of the PCK metric is that it is only computed between image pairs. 
However, the errors in keypoint prediction tend to accumulate when transferring keypoints over a sequence of images. To address this limitation, we propose a new metric to measure consistency across multiple images, called $\kcycle$. 
Given a set of $k$ images $\{\I^1, \I^2, \dots, \I^k\}$ and an annotated keypoint in the first image $\p^1$ visible in \textbf{each} of the $k$ images, we propagate $\p$ from $\I^1 \rightarrow \I^2, \I^2 \rightarrow \I^3,\dots, \I^{k-1} \rightarrow \I^{k}, \I^{k} \rightarrow \I^{1}$ and get the corresponding predictions $\p^{1\rightarrow2}, \p^{1\rightarrow2\rightarrow3},\dots$ and so on. As before, $\p^{1\rightarrow \dots \rightarrow j}$ is considered to be predicted correctly if it is within a threshold $\alphapck\cdot\max{(H_\bbox, W_\bbox)}$ of the ground truth keypoints $\hat{\p}^{1\rightarrow \dots \rightarrow j}$. We sum up all the correct predictions and plot scores at different values of $\alphapck$ in \cref{fig:cycle_pck}. 
We choose $k=4$ for all experiments (with additional results for other values of $k$ provided in the supplement). Note that since the number of possible permutations of $k$-length sequences can be very large, we randomly sample 200 sequences in our experiments.

\cref{fig:cycle_pck} shows that our method significantly outperforms the DINO+NN baseline for both small and large values of $\alphapck$ across all datasets
(complete results in the supplemental material). We attribute this result to having a consistent canonical space across the image collection that 
prevents errors in keypoint transfer from accumulating to large values.

\subsection{Ablations}
\label{sec:ablation}
We perform an ablation study on our various proposed losses proposed, summarized in \cref{tab:ablation}. We report average PCK@$0.10$ results for first 3 categories of the CUB-200 dataset and all 4 categories of the PF-Willow dataset. As expected, the keypoint loss $\mathcal{L}_\text{KP}$ plays the most important role in our overall framework. We also found the total variation regularization $\mathcal{L}_\text{TV}$ to be crucial for network training convergence. $\mathcal{L}_\text{Equi}$ is necessary for learning dense correspondence. Finally, $\mathcal{L}_\text{Recon}$ and $\mathcal{L}_\text{Parts}$ provide comparatively small improvements.


\begin{table}[!ht]
\centering
    \caption{\textbf{Ablation Study.} Average PCK@$0.10$ (for 3 and 4 categories respectively) in CUB and PF-Willow datasets.}
    \vspace{-1em}
    \label{tab:ablation}
    \resizebox{0.9\linewidth}{!}{
    \begin{tabular}{ lcc } 
    \toprule
    \textbf{Ablation} & CUB-200 & PF-Willow \\
                      & (3 categories) & (4 categories) \\
    \midrule
    Complete objective & 75.9 & 76.3 \\
    No  $\mathcal{L}_\text{KP}$     & 22.8 & 36.2 \\
    No  $\mathcal{L}_\text{TV}$     & 43.9 & 40.4 \\
    No  $\mathcal{L}_\text{Equi}$   & 64.8 & 65.6 \\
    No  $\mathcal{L}_\text{Recon}$  & 73.3 & 74.2 \\
    No  $\mathcal{L}_\text{Parts}$  & 73.6 & 73.5 \\
    \bottomrule
    \end{tabular}
    }
\end{table}

\begin{figure}[!ht]
 \centering
 \includegraphics[width=\linewidth,trim={0 0 0 0}, clip]{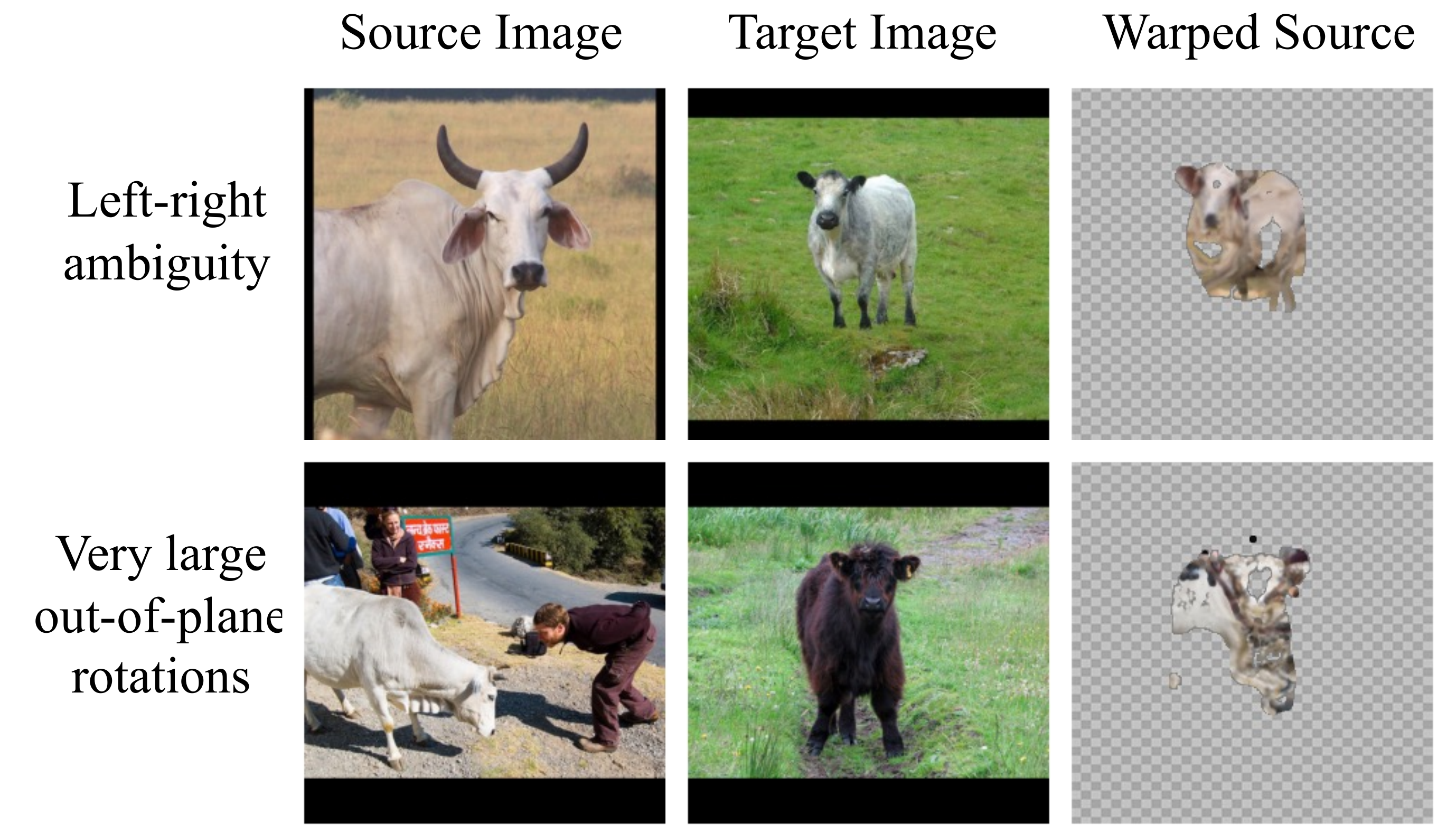}
 \vspace{-1em}
 \caption{\footnotesize \textbf{Limitations.} Top row shows that our model can map left part of object in source to right part of object in target when object is symmetric. Bottom row shows that our model fails for very large out-of-plane rotations.}
 \label{fig:challenges}
 \vspace{-1em}
\end{figure}
\subsection{Limitations}
\noindent\textbf{Left-right ambiguity:}
One shortcoming of our approach is that it cannot differentiate well between left and right parts well for symmetric objects. We attribute this problem to the SSL models being invariant to left-right flips during their training. The top row of \cref{fig:challenges} shows that our model matches the left part of the cow torso in the source image to the right part of the torso in the target image (note left part of cow is not visible in the target image). Some heuristics used in prior works, such as flipping the source and target images and picking a combination that provides minimum total variation loss, could be used in our work as well. For brevity, we provide results without this heuristic.
\noindent\textbf{Large shape changes:}
Our model doesn't handle large viewpoint changes well, especially when there are few intermediate viewpoints. In the bottom row of \cref{fig:challenges}, we see that model is unable to warp the source cow image to target image even for the co-visible portions.

\section{Conclusions}
\label{sec:conclusion}

We propose \backronym, a method to address the challenging task of dense correspondences across images of an object or object category captured in-the-wild. \backronym utilizes noisy and sparse pseudo-correspondences in pre-trained ViT feature space to build an accurate and dense consistent mapping from image to a canonical space. Extensive qualitative and quantitative experiments show that \backronym works in low-shot settings and can deal with extreme variations in pose, background, occlusion, and object deformations.
We also propose a new metric $\kcycle$ to evaluate the consistency of keypoint predictions over a set of images beyond pair-wise consistency.
\backronym can obtain consistent dense mappings competitive with supervised counterparts with just a few images. In future work, we will explore applications of \backronym in other few-shot downstream tasks such as reconstruction, pose estimation and tracking.

\clearpage
{\small
\bibliographystyle{ieee_fullname}
\bibliography{egbib}
}

\clearpage
\appendix
\section{Implementation Details}

\subsection{$\A$ Architecture}

We use a small U-Net architecture~\citep{ronneberger2015u} to represent $\A$ consisting of four downscaling fully convolutional blocks and four upscaling fully convolutional blocks. Output of downscaling blocks is concatenated to the upscaling blocks, as is typical in U-Nets. The size and parameter details of each of the blocks is provided in \Cref{tab:unet}.

Output of the final layer has two channels predicting the $x$ and $y$ canonical space coordinates for each pixel. The canonical grid is a learned embedding of dimension $256\times256\times4$. We fixed the learning rate for the network to $0.001$ and train the entire network end to end for $20,000$ iterations with a batch size of 20 on a single GPU.

\begin{table}[!htb]
    \centering
    \footnotesize
    \caption{U-Net architecture for $\A$ - It is a fully convolutional architecture, consisting of an Input block, Output block, four Up blocks, and four Down blocks. Each block consists of Up, Down, and DoubleConv Layers as shown below. Each DoubleConv, Up, and Down blocks is parameterized by number of input and output channels \ie, $(C_{in}, C_{out})$. Each Conv2D is represented by $(C_{in}, C_{out}, \text{kernel}, \text{stride}, \text{pad})$. BN stands for batch normalization layer and has $C_{out}$ parameters. ReLU are Rectified Linear Units without any parameters.}
    \begin{tabular}{@{}l|lc@{}}
    \toprule
    Blocks & Layers & Output Size\\
    \midrule
    
    Input  & DoubleConv $(3, 32)$    & $32 \times 128 \times 128$  \\
    Down-1 & Down       $(32, 64)$   & $64 \times 64 \times 64$  \\
    Down-2 & Down       $(64, 128)$  & $128 \times 32 \times 32$ \\
    Down-3 & Down       $(128, 256)$ & $256 \times 16 \times 16$  \\
    Down-4 & Down       $(256, 512)$ & $512 \times 8 \times 8$  \\
    Up-1   & Up         $(512, 128)$ & $256 \times 16 \times 16$ \\
    Up-2   & Up         $(256, 64)$ & $128 \times 32 \times 32$  \\
    Up-3   & Up         $(128, 32)$  & $64 \times 64 \times 64$ \\
    Up-4   & Up         $(64, 32)$   & $32 \times 128 \times 128$\\
    Output & DoubleConv $(32, 4)$    & $4 \times 256 \times 256$ \\
    \midrule
    DoubleConv & $\begin{array}{@{}l}  \text{Conv2D}(C_{in}, C_{out}, 3, 1, 1) \\ \text{BN}(C_{out}) \\ \text{ReLU}  \\   \text{Conv2D}(C_{in}, C_{out}, 3, 1, 1)\\ \text{BN}(C_{out}) \\ \text{ReLU} \end{array}$ &\\
    \midrule
    Down & $\begin{array}{@{}l}  \text{MaxPool2D}(2) \\ \text{DoubleConv}(C_{in}, C_{out}) \end{array}$ &\\
    \midrule
    Up & $\begin{array}{@{}l}  \text{BilinearUpsample}(2) \\ \text{DoubleConv}(C_{in}, C_{out}) \end{array}$ &\\
    \bottomrule
    \end{tabular}
    \label{tab:unet}
\end{table}

\subsection{Canonical grid $\G$}
The canonical grid $\G$ consists of a simple $256 \times 256\times 4$ feature grid which is learned during the training with the same learning rate and optimizer as the alignment network $\A$. Each location in $\G$ stores an $(r,g,b,\alpha)$ value which corresponds to colors $(r,g,b)$ and a probability $\alpha$ that this location corresponds to a foreground pixel in the image.

\subsection{Loss terms}

Recall that our overall objective function comprises 5 different loss terms of which $\loss_\text{KP}$, $\loss_\text{Equi}$, and $\loss_\text{TV}$ are applied to canonical space coordinates $\C$ and update only the parameters of alignment network $\A$ (and not the canonical grid $\G$). $\loss_\text{Recon}$ and $\loss_\text{Parts}$ can backpropagate gradients to both the alignment network and the canonical grid. In all our experiments, on all 4 datasets and their respective categories, we use the same set of weight coefficients (except for the ablation study in Section 4, where we make the coefficients zero one at a time). We set of the coefficients for different loss terms as following: $\lambda_\text{KP}=10$, $\lambda_\text{Equi}=1$, $\lambda_\text{TV}=9000$, $\lambda_\text{Recon}=1$, and $\lambda_\text{Parts}=10$. We observed that our framework is robust to the choice of hyperparameters. We can further increase the PCK performance by setting per-category hyperparameters, however, per-category (or per-collection) tuning is not ideal for scaling the model to a large number of image collections. Hence, we choose to report all our numbers with a fixed set of hyperparameters. 

\subsection{Choice of SSL for pseudo-correspondences.} In our experiments, we obtain initial set of pseudo-correspondences by finding mutual nearest neighbors from frozen DINO (ViT-S/8) network. Note that DINO is not trained or fine-tuned in our experiments. Our alignment network $\A$, which is much smaller than DINO, is trained from scratch. This is also in contrast with other weakly supervised techniques such as PMD which uses ResNet-101 / VGG-16 ($> 40\text{M}$ params). We observe that performance of our framework can be improved further by using better pseudo-correspondences. \cref{tab:ablation_bb} shows ASIC results when obtaining pseudo-correspondences from 3 different ViT architectures.

\begin{table}[!ht]
\centering
    \caption{Pseudo-correspondences Ablation on CUB-001}
    \label{tab:ablation_bb}
    \resizebox{\linewidth}{!}{
    \begin{tabular}{ lcc|cc } 
    \toprule
    \multirow{2}{*}{\textbf{Architecture}} & \multirow{2}{*}{\# params} & ImageNet Top-1 & DVD & ASIC (ours) \\
    &  & (Accuracy) & PCK@$0.1$ & PCK@$0.1$  \\
    \midrule
    ViT-S/16        & 21M & 77.0 & 59.8 & \textbf{63.7} \\
    ViT-B/8         & 85M & 80.1 & 66.4 & \textbf{74.9} \\
    ViT-S/8 (paper) & 21M & 79.7 & 66.8 & \textbf{71.8} \\
    \bottomrule
    \end{tabular}
    }
\end{table}

\section{Visualizing the Canonical Space}
Recall that in Section 4 of the paper, we showed the canonical space mapping for various datasets learned by our model.
Here we provide further details of the canonical space mapping.
Specifically we first show the region in 2D space where each point in the image is getting mapped in \Cref{subsec:colormap}.
Next we show the RGB grid that is learned by our model.

\begin{figure}[!ht]
 \centering
 \includegraphics[width=\linewidth,trim={0 0 0 0}, clip]{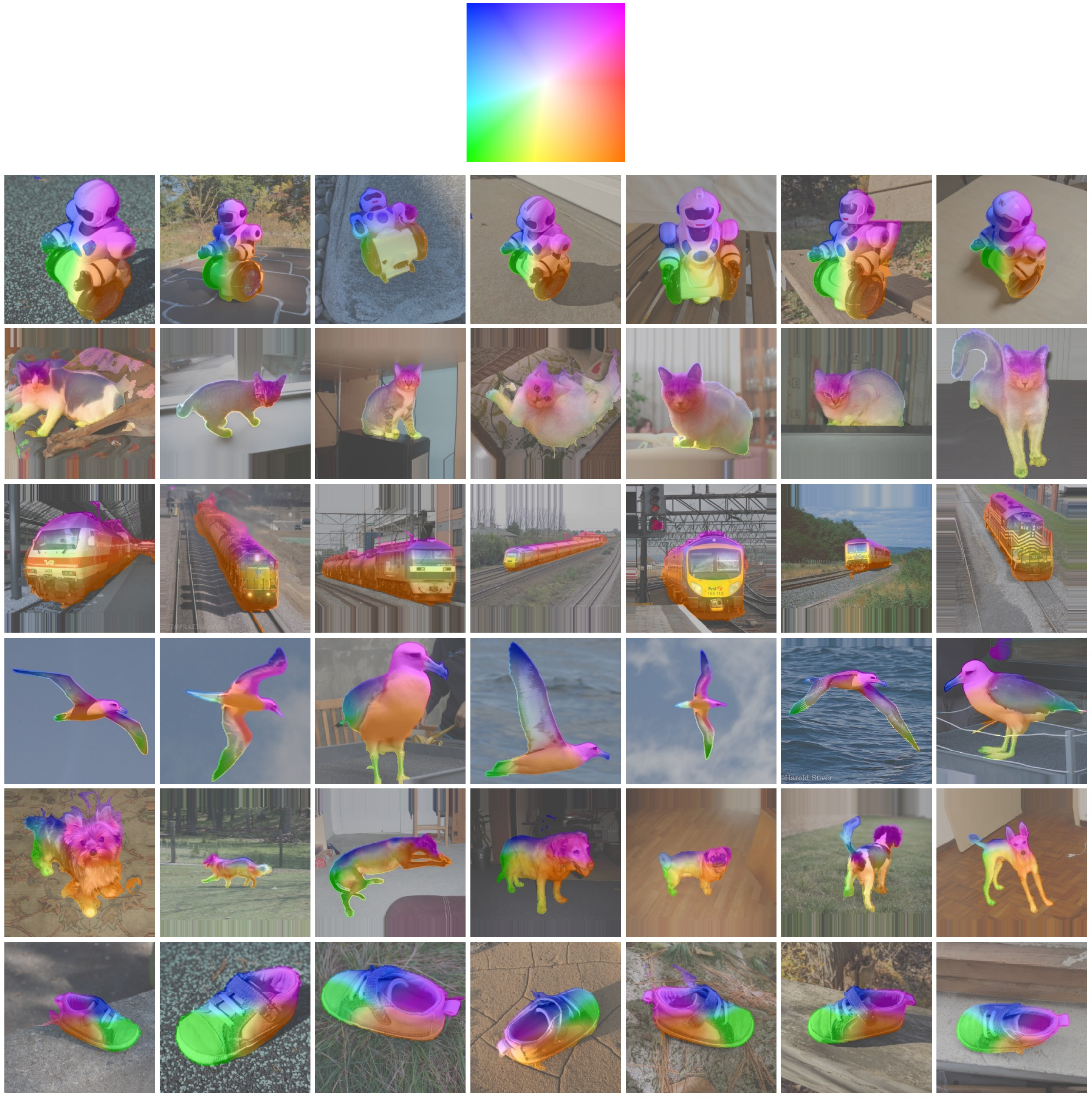}
 \caption{\footnotesize \textbf{Colormap for canonical space visualization.} We use the colormap shown in the top row to represent the canonical space. Based on the canonical space coordinates predicted by our model for each pixel, we copy (or more precisely splat) the colors from the canonical space colormap to the original image. Each row shows the mapping learned by the model for different datasets.}
 \label{fig:colormap}
\end{figure}

\subsection{With colormap}
\label{subsec:colormap}
First, we reproduce the results from Section 4 here, along with the colormap of canonical space used to visualize them in \Cref{fig:colormap}. Note that we train a different model for each dataset. The figure shows that the semantically similar parts of objects get mapped to nearby location in the canonical space. Our model is able to learn a smooth mapping for each object.

\subsection{With learned RGB Grid}
\begin{figure}[!ht]
 \centering
 \includegraphics[width=\linewidth,trim={0 0 0 0}, clip]{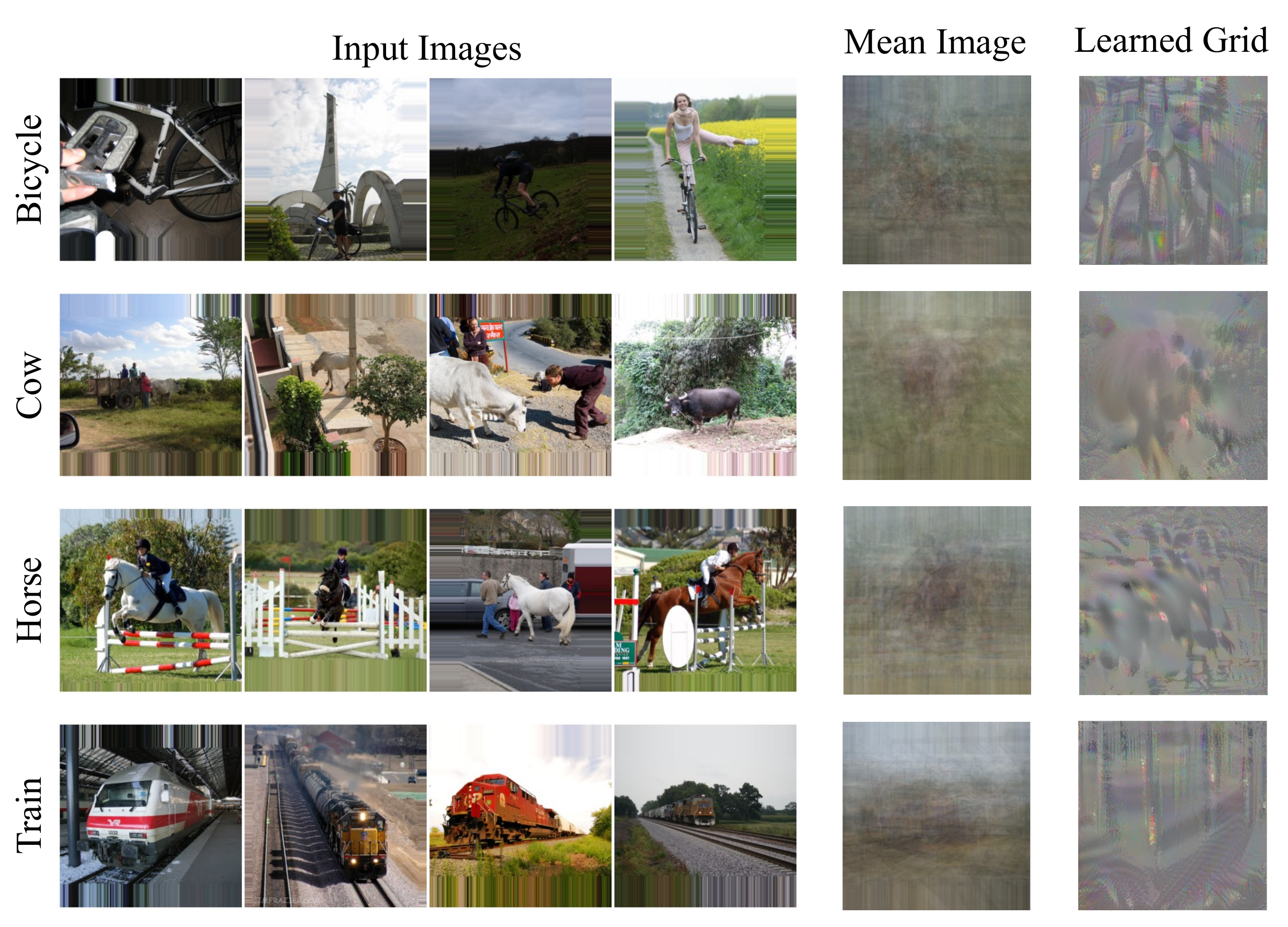}
 \caption{Sample images from the dataset in the left four columns, followed by the mean image of the dataset. The last column shows the joint canonical grid learned by the model.}
 \label{fig:grid}
\end{figure}

Our method also learns an RGB grid. \Cref{fig:grid} shows the grid learned for 4 different datasets. We observe that while our grid is not interpretable, there are distinct patterns that emerge for each dataset. Specifically, one can observe wheel-like shapes in the bicycle grid, and a cube in the train canonical grid. We attribute the weak interpretability of the learned grid to the large variability in the challenging in-the-wild images, where images may consist of different instances of an object category in very different poses, articulations, shapes, textures, background, and lighting. The collections we used are also very small (5-25 images). Further while our alignment network ensures that the pseudo-correspondences across images land at the same location in the canonical space, nearby points within the same image can still map to far away locations in the canonical space.
Making the grid more interpretable could be useful for better understanding of the model's capabilities and limitations. For instance, in the case of GANgealing~\cite{peebles2022gan}, training a GAN on a large dataset of cats ($\sim$1.5M images), they are able to learn a canonical atlas which looks like face of a cat. This allows them to use canonical atlas as the template for image editing and edit propagation templates (although, a limitation of this approach is that it doesn't allow editing any parts other than the face of a cat).

\section{Results on different k-values and all datasets for $\kcycle$}

\begin{figure*}[!ht]
 \centering
 \begin{minipage}{\textwidth} 
 \includegraphics[width=\linewidth,trim={0 0 0 0}, clip]{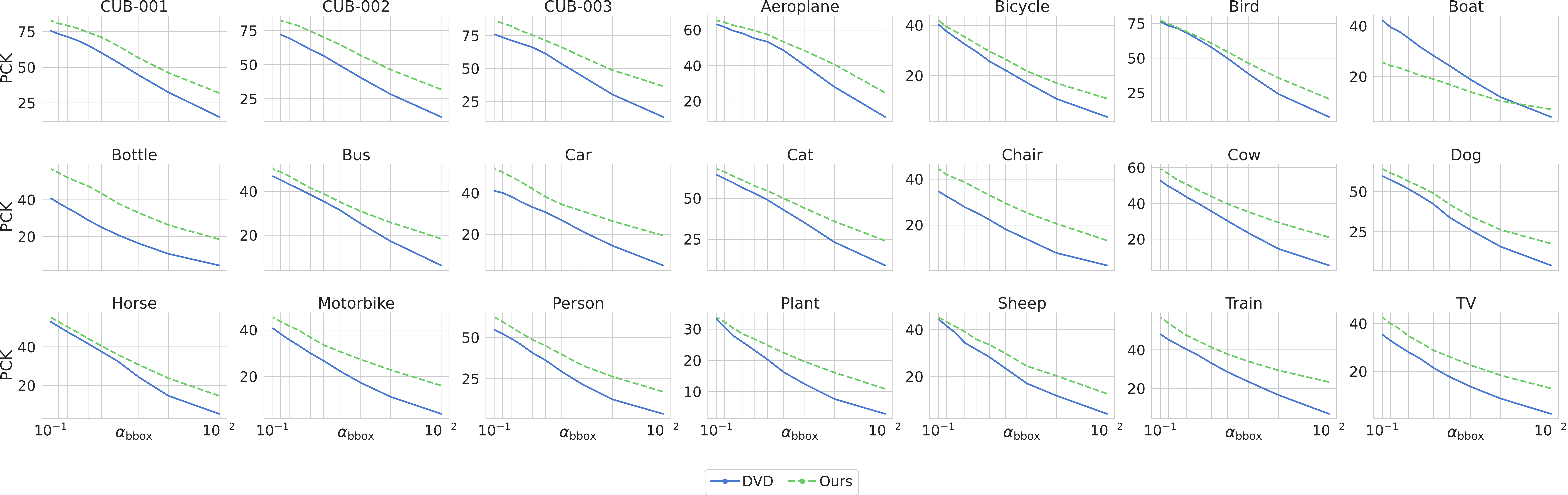}
 \caption{$\mathbf{2}\cycle$ for three CUB-200 categories and 18 SPair-71k categories (test split)}
 \label{fig:2cycle}
 \vspace{3em}
 \end{minipage}

\begin{minipage}{\textwidth} 
\includegraphics[width=\linewidth,trim={0 0 0 0}, clip]{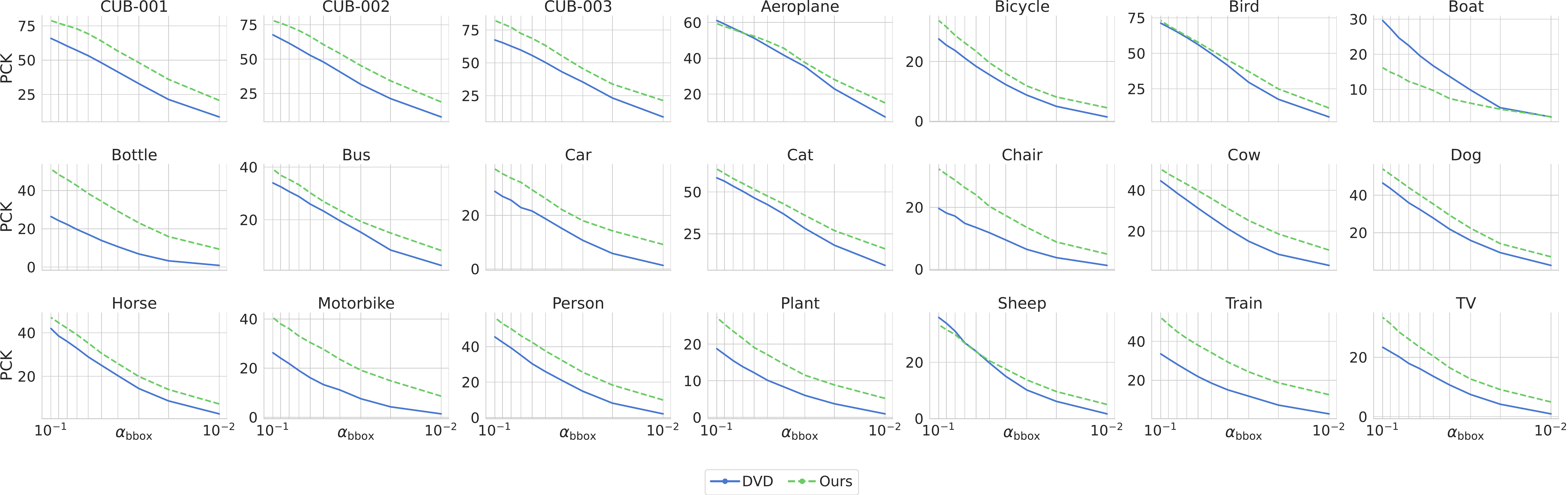}
 \caption{ $\mathbf{3}\cycle$ for three CUB-200 categories and 18 SPair-71k categories (test split) }
 \label{fig:3cycle}
 \vspace{3em}
\end{minipage}

\begin{minipage}{\textwidth} 
 \centering
 \includegraphics[width=\linewidth,trim={0 0 0 0}, clip]{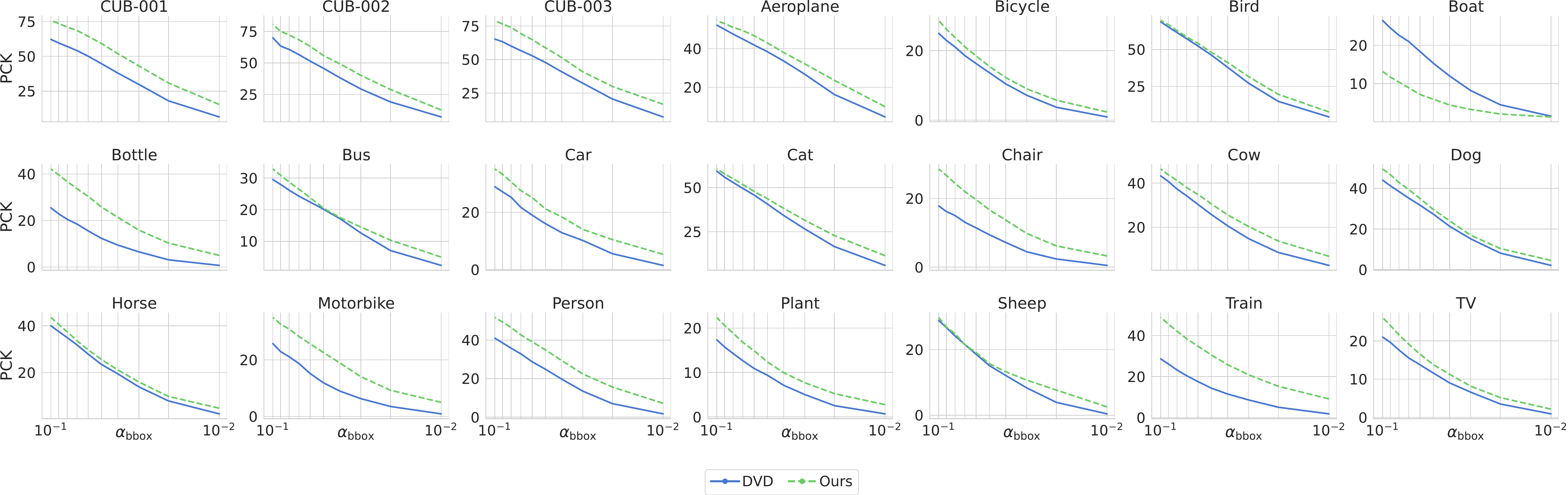}
 \caption{$\mathbf{4}\cycle$ for three CUB-200 categories and 18 SPair-71k categories (test split)}
 \label{fig:4cycle}
 \vspace{3em}
 \end{minipage}
\end{figure*}

We share the results for of $k \in \{2,3,4\}$ and plot $\kcycle$ for all the datasets (with groundtruth keypoint annotations) we considered in our experiments. \Cref{fig:2cycle,fig:3cycle,fig:4cycle} show the comparison between our method and DVD. Note that DVD is also referred to as DINO + NN (where NN stands for nearest neighbors) in the main paper to clarify the strategy used to find the correspondences. Our method consistently outperforms the baseline, at both small and large values of $\alphapck$ (which corresponds to the coarse and fine precision or accuracy of the transfer).

\end{document}